\documentclass{article}

\usepackage[nonatbib, final]{neurips_2021}

\usepackage[utf8]{inputenc} 
\usepackage[T1]{fontenc}    
\usepackage[numbers]{natbib}
\usepackage{hyperref}       
\usepackage{url}            
\usepackage{booktabs}       
\usepackage{amsfonts}       
\usepackage{nicefrac}       
\usepackage{microtype}      
\usepackage{xcolor}         
\usepackage{graphicx}
\usepackage{multirow}

\title{Transformer-based Flood Scene Segmentation for Developing Countries}

\author{%
   Ahan M R\thanks{Equal Contribution}\\
   BITS Pilani Goa Campus \\
  \texttt{ahanmr98@gmail.com} \\
  \And
  Roshan Roy\textsuperscript{*} \\
  BITS Pilani\\
  \texttt{rroshanroy@gmail.com} \\
   \And
   Shreyas Sunil Kulkarni\\
   BITS Pilani Hyderabad Campus \\
  \texttt{sskshreyas@gmail.com} \\
    \And
   Vaibhav Soni \\
   MANIT Bhopal \\
  \texttt{vaibsoni@gmail.com} \\
  \And
   Ashish Chittora \\
   BITS Pilani Goa Campus \\
  \texttt{ashishc@goa.bits-pilani.ac.in}
}

\begin{document}
\maketitle

\begin{abstract}
Floods are large-scale natural disasters that often induce a massive number of deaths, extensive material damage, and economic turmoil. The effects are more extensive and longer-lasting in high-population and low-resource developing countries. Early Warning Systems (EWS) constantly assess water levels and other factors to forecast floods, to help minimize damage. Post-disaster, disaster response teams undertake a Post Disaster Needs Assessment (PDSA) to assess structural damage and determine optimal strategies to respond to highly affected neighborhoods. However, even today in developing countries, EWS and PDSA analysis of large volumes of image and video data is largely a manual process undertaken by first responders and volunteers. We propose \textit{FloodTransformer}, which to the best of our knowledge, is the first visual transformer-based model to detect and segment flooded areas from aerial images at disaster sites. We also propose a custom metric, \textit{Flood Capacity (FC)} to measure the spatial extent of water coverage and quantify the segmented flooded area for EWS and PDSA analyses. We use the SWOC Flood segmentation dataset and achieve 0.93 mIoU, outperforming all other methods. We further show the robustness of this approach by validating across unseen flood images from other flood data sources.
\end{abstract}

\section{Introduction and Context}
The Center for Research on the Epidemiology of Disasters, in affiliation with the World Health Organization (WHO), reported that natural disasters accounted for 1.3 million deaths and over USD 2 trillion in economic damage — all between 1998 and 2017 \cite{death-stats}. Flooding related damage is a factor in most of them \cite{flood-major} and frequent the list of most expensive disasters \cite{top-disaster-stats}. Developing economies of Asia are disproportionately affected and are the worst-hit by floods, accounting for 44\% of all flood disasters from 1987-1997 \cite{asia-stats}. India alone registers 1/5th of global deaths from floods \cite{india-stats}. 
Rapid urbanization, global climate change, and rising sea water levels will expose 1.47 billion more people to flood risk, with 89\% of them living in low-middle income countries \cite{projections}. Flood Segmentation technology is instrumental for Disaster Prediction and Response is critical to save lives and livelihoods. 

\textbf{Flood Response}: Typically, disaster management teams complete a Post Disaster Needs Assessment (PDSA) and rapidly develop infrastructure based on this report on the collected data \cite{FEMA2016}. Unmanned Aerial Vehicles (UAVs) are deployed to collect large volumes of image and video data in affected regions. PDSA is essential for identification of submerged regions, sanity check of large building structures, debris identification, and search-and-rescue (S\&R) operations.

\textbf{Flood Forecasting}: Flood segmentation techniques can be critically important for flooding-related Early Warning Systems (EWS). According to research in \cite{flood-monitoring}, Indians given a flood warning are twice as likely to evacuate safely than Indians without any notice. which require constant monitoring of river or sea water levels. Comparison of current levels with historical evidence of flood-prone water levels can help understand when to trigger warnings appropriately. 

\textbf{Constraints}: Developing countries are plagued by resource and economic constraints. Failure of macro- and micro- infrastructure planning in Nicaragua led to re-construction on top of an earthquake faultline \cite{drawbacks}. Weak social safety and insurance policies inflate recovery time \cite{developing}.
Economic vulnerability renders countries like Haiti, Ethiopia, Nepal, El Salvador in a near-permanent state of emergency alert \cite{drawbacks}. In these countries, processing and analysis of large-scale visual data from UAVs for PDSA in Flood response is a manual process that requires multi-team intervention, which poses a serious bottleneck in search-and-response speed. Deployment of EWSs is infeasible because human monitoring of video feeds is too cumbersome and expensive.

\textbf{AI Technology}: To reduce the burden of manual analysis on crisis responders, Deep learning is well-suited to scale, automate and expedite these operations. 
The last few years have witnessed a tremendous rise in CNN-based image classification and segmentation research \cite{dl_survey}. However, CNNs suffer from a well-known problem — large inductive biases. Conceretely, CNNs assume locality and translation equivariance, which hurt the interpretability of pure CNN-based algorithms. Recently, visual transformers have garnered attention for image classification, segmentation and object detection tasks \cite{tf_class, Xie2021SegFormer,zhu2021deformable, transformers1} for challenging these assumptions with comparable accuracy.

\textbf{Contributions}: In this work, we propose a hybrid fused CNN-Transformer: \textit{FloodTransformer} to tackle flood water segmentation on the Water Segmentation Open Collection (WSOC) dataset \cite{wsoc}. First, we achieve state-of-the-art results and are the first work (to the best of our knowledge) to apply new transformer-driven research to the flood data domain. Second, our approach is extendable — we demonstrate the ability of our model to generalize well on unseen data sources. Further local calibration, if required at all, simply requires weight fine-tuning with previous, region-specific, flood scene data. Third, our model does not suffer from data scarcity - it only requires image data input and not complex sensor data which is hard to collect \cite{forecasting}. Last, the transformer-based encoder applies recent DL innovations to the flood data domain. Although the hybrid method still uses CNNs in the decoder network, the aforementioned spatial inductive biases no longer occur throughout the entire network. Dependencies between patch embeddings are learnt from scratch. This improves the robustness of our approach. 

\section{Methodology}
To achieve the Flood Scene Understanding, we introduce a Deep Learning model for Flood image segmentation and quantify the impact of flooding with a custom metric called \textit{Flooding Capacity}.
\label{gen_inst}

\subsection{Method}
\label{method}
\begin{figure}[h]
    \centering
    {{\includegraphics[width=14cm]{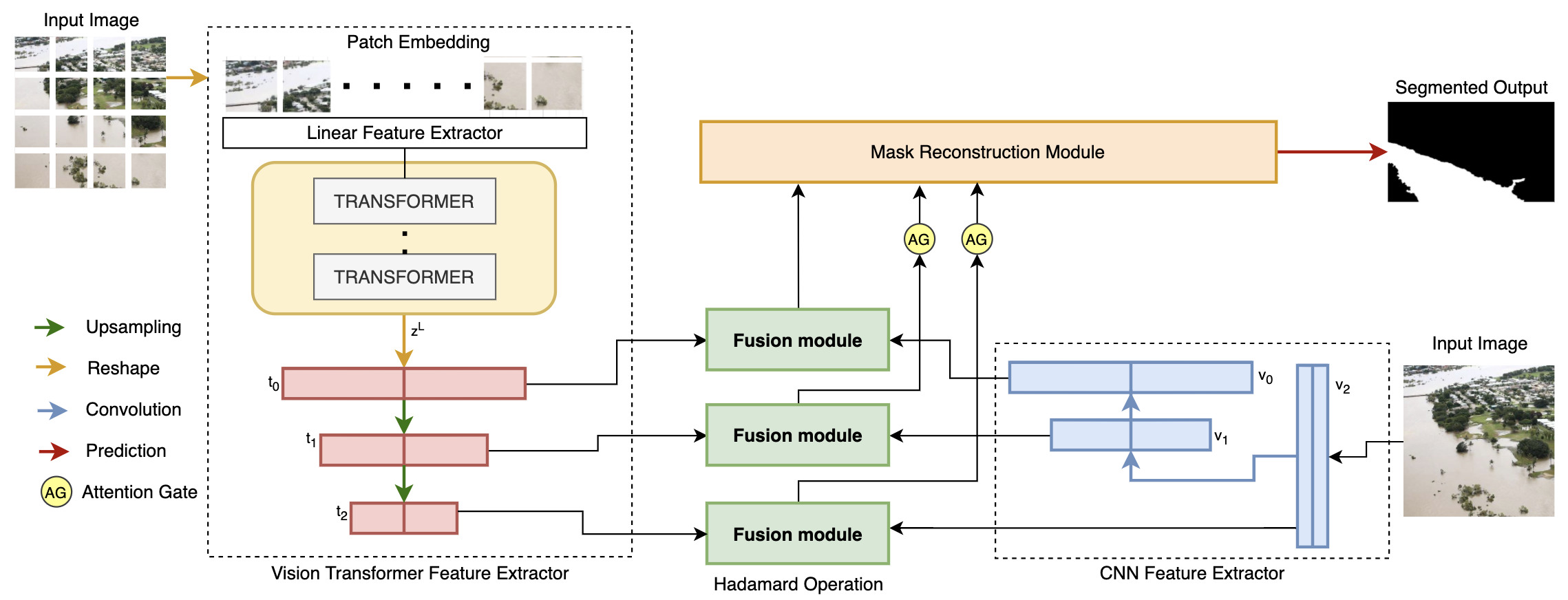}}}%
    \enskip
    \caption{\textit{FloodTransformer} architecture: simplified from Zhang \textit{et al.} \cite{zhang2021transfuse} for representation.}%
    \label{fig:model-architecture}%
\end{figure}

Inspired by Zhang \textit{et al}. \cite{zhang2021transfuse}, we propose \textit{FloodTransformer} to solve segmentation for the flood data domain. It is a fusion architecture of Visual Transformer \cite{zhou2021deepvit} and Convolution Neural Networks (CNNs) and its model architecture is displayed in Figure \ref{fig:model-architecture}.  

Complex flooding imagery may contain heterogeneous objects, flooding patterns and backgrounds. Using the self-attention module of the visual Transformer module from  \cite{zhou2021deepvit} and global vector representation learned from the CNN network, \textit{FloodTransformer} fuses the trained embeddings to learn long-term spatial relationships between the aforementioned entities in images of flood affected areas. 
Using Hadamard bilinear product \cite{zhang2021transfuse}, the fusion module fuses information via embeddings from both parallel streams into a dense representation. The combination of multi-level fusion maps generates the segmentation output of the model. We summarize each component below, per Zhang \textit{et al.} \cite{zhang2021transfuse}. 

\textbf{Transformer Module}:
We use the encoder-decoder network using Visual Transformer \cite{zhou2021deepvit}. The input image $\mathbf{x} \in \mathbb{R}^{H \times W \times 3}$ is sliced into N patches, where $\mathbf{N} = {\frac{H}{F} \times \frac{W}{F} }$ and F is usually set to 16 or 32. These patches are flattened into a linear projection layer to generate an image embedding, added to a trainable position embedding, and passed into $L$ layers of the Transformer encoder's Multi-head Self Attention (MSA) mechanism. Every layer updates the embedding with $[q, k, v]$ triplets and the projection matrix. The encoded sequence $z^{L}$ is passed to the Transformer decoder, which aims to recover spatial context and outputs feature maps $t^{0}$, $t^{1}$, $t^{2}$ of different scales.

\textbf{CNN Module}: Per \cite{zhang2021transfuse} and \cite{roytowards}, we implement three Res-blocks as a shallow CNN network; Transformers already provide global context and retain rich local information and relationships. In parallel to the transformer module, the three CNN blocks output downsampled feature maps $v_{0}$, $ v_{1}$, $ v_{2}$ of different scales.  

\textbf{Fusion Module}:
Outputs $ v_{0},t_{0} \in \frac{H}{8} \times \frac{W}{8}$ , $v_{1}, t_{1} \in \frac{H}{4} \times \frac{W}{4}$ and $v_{2}, t_{2} \in \frac{H}{2} \times \frac{W}{2}$ from both modules are fused via the Fusion module from Zhang \textit {et al}. \cite{zhang2021transfuse}. The fused vector $f_{i}$ is generated by implementing the following operations: 

\begin{figure}[ht]
    \centering

    {{\includegraphics[width=12cm]{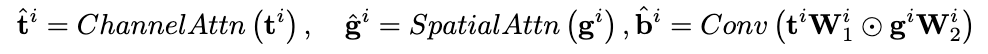}}}%
    \enskip
    \label{fig:addImage}%
\end{figure}


where $\odot$ is the Hadamard product as described in \cite{zhang2021transfuse}.  

\subsection{Metric}
We introduce a custom metric called \textit{Flooding Capacity (FC)} to quantify the impact of flooding in an image. It is the percentage of flooding in an image, and is calculated from the segmented mask output of the image, as described in Equation \ref{eqn}. 

\begin{equation}
Flooding Capacity (FC) =\frac{1}{\sum_{i=0}^{k} \sum_{j=0}^{k} (p_{i j}==0) || (p_{ij}==1)} \sum_{i=0}^{k} \sum_{j=0}^{k} (p_{i j}==1)
\label{eqn}
\end{equation}

$ p_{ij}$ are the pixel values of the binary mask output and can have values ${0,1}$. We calculate \verb|Flooding area| as the sum of all $ p_{ij} == 1$ \verb|flooding| pixels. We divide \verb|flooding area| by \verb|total area|, which is the sum of all $ p_{ij} == 0 $ \verb|non-flooding| and $ p_{ij} == 1 $ \verb|flooding| pixels. 

Flood Scene Segmentation, as described in Section \ref{method}, used with \textit{Flood Capacity} calculation from Equation \ref{eqn} can be used on images pre/post flooding for complete Scene understanding. For Flood Response, this helps optimize operational planning. For Flood Forecasting, this can quantify the water-cap in a particular region over time separated image inputs.

\section{Dataset Description}
\textbf{Water Segmentation Open Collection (WSOC)} dataset \cite{wsoc} consists of 11900 flood or water body related images, composed from multiple open-source source datasets such as COCO \cite{coco}, Semantic Drone Dataset \cite{semantic-drone} and other datasets containing flood-related images \cite{wsoc}. These images were shot from a variety of sources such as fixed surveillance cameras, drones (UAVs), crowdsourced in-field observations (on foot, boat, vehicle) and from social media streams. Some sample images are shown in Figure \ref{fig:WSOC}. 

Image resolutions vary from 147x150 pixels for low-quality social media streams to 2448x3264 pixels for sophisticated aerial imagery. 
Sampling from disparate sources is the reason for significant variations in visual appearance, illumination, water body, and geography; this makes water segmentation challenging. Newly-added social media images undergo a three-step manual annotation process to generate binary ground truth masks for pixel-wise classification between water or no-water. 

\section{Results}

\begin{figure}[ht]
    \centering

    {{\includegraphics[width=14cm]{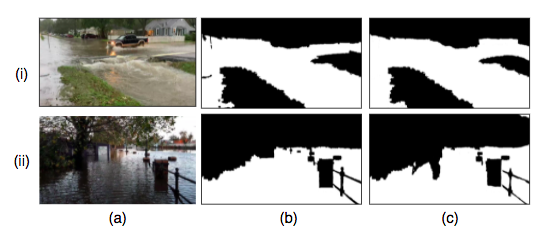}}}%
    \enskip
    \caption{\centering Qualitative results on (i) social media (ii) in-field images in WSOC dataset. \newline \centering (a) Image (b) Ground Truth (c) Segmented Output}%
    \label{fig:WSOC}%
\end{figure}

We follow the WSOC method \cite{wsoc} and perform a 90-10 train-test split of the data. Qualitative results of output segmentation masks are shown in Figure \ref{fig:WSOC}. We evaluate the performance of our hybrid Flood Transformer model with mean intersection-over-union (mIOU) and Percentage Accuracy (PA) scores, on WSOC, as shown in Table \ref{tab:my-table-1}. We compare our result to that of other benchmark models on WSOC \cite{wsoc}, in Table \ref{tab:my-table-1}. 

As shown, \textit{FloodTransformer} comfortably achieves state-of-the-art segmentation mIOU of 0.93. It is based on Zheng et al \cite{zhang2021transfuse}'s \textit{TransFuse} architecture, which comfortably beats traditional benchmarks such as UNet \cite{unet}. The Flood Capacity (FC) from Equation \ref{eqn} of predicted mask in Fig \ref{fig:WSOC}(i)(c) is 0.53 and of mask in Fig \ref{fig:WSOC}(ii)(c) 0.47. 

\begin{figure}[ht]
    \centering

    {{\includegraphics[width=10cm]{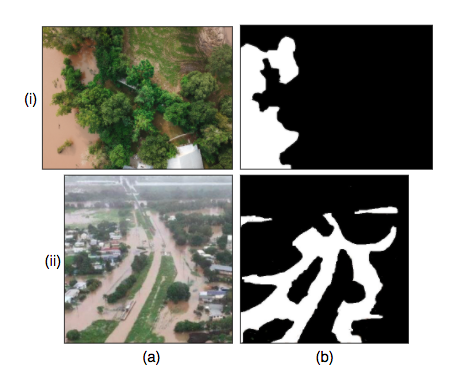}}}%
    \enskip
    \caption{\centering Qualitative results on out-of-dataset, aerial flooding imagery. \newline \centering (a) Image (b) Segmented Output}%
    \label{fig:Unlabelled}%
\end{figure}

To demonstrate the robustness of our model with further qualitative results, we perform inference on out-of-dataset images, as shown in Figure \ref{fig:Unlabelled}. The Flood Capacity (FC) from Equation \ref{eqn} for the predicted mask in Fig \ref{fig:Unlabelled}(i)(b) is 0.18 and for mask in Fig \ref{fig:Unlabelled}(ii)(b) is 0.26. 

\begin{table*}[]
\centering
\begin{tabular}{llllll}
\hline
\multicolumn{1}{|l|}{\multirow{3}{*}{\textbf{Model}}} &
  \multicolumn{1}{l|}{\multirow{3}{*}{\textbf{Backbone}}} &
  \multicolumn{4}{l|}{\textbf{WSOC Dataset}} \\ \cline{3-6} 
\multicolumn{1}{|l|}{} &
  \multicolumn{1}{l|}{} &
  \multicolumn{2}{l|}{\textbf{mIOU {[}\%{]}}} &
  \multicolumn{2}{l|}{\textbf{PA {[}\%{]}}} \\ \cline{3-6} 
\multicolumn{1}{|l|}{} &
  \multicolumn{1}{l|}{} &
  \multicolumn{1}{l|}{$\mu$} &
  \multicolumn{1}{l|}{$\sigma$} &
  \multicolumn{1}{l|}{$\mu$} &
  \multicolumn{1}{l|}{$\sigma$} \\ \hline
\multicolumn{1}{|l|}{Tiramisu \cite{tiramisu}} &
  \multicolumn{1}{l|}{None} &
  \multicolumn{1}{l|}{0.38} &
  \multicolumn{1}{l|}{0.17} &
  \multicolumn{1}{l|}{0.73} &
  \multicolumn{1}{l|}{0.24} \\ \hline
\multicolumn{1}{|l|}{SegNet \cite{segnet}} &
  \multicolumn{1}{l|}{ResNet50} &
  \multicolumn{1}{l|}{0.85} &
  \multicolumn{1}{l|}{0.08} &
  \multicolumn{1}{l|}{0.94} &
  \multicolumn{1}{l|}{0.01} \\ \hline
\multicolumn{1}{|l|}{UNet \cite{unet}} &
  \multicolumn{1}{l|}{ResNet50} &
  \multicolumn{1}{l|}{0.88} &
  \multicolumn{1}{l|}{0.06} &
  \multicolumn{1}{l|}{0.94} &
  \multicolumn{1}{l|}{0.03} \\ \hline
\multicolumn{1}{|l|}{FCN32 \cite{fcn32}} &
  \multicolumn{1}{l|}{ResNet50} &
  \multicolumn{1}{l|}{0.79} &
  \multicolumn{1}{l|}{0.10} &
  \multicolumn{1}{l|}{0.87} &
  \multicolumn{1}{l|}{0.08} \\ \hline
\multicolumn{1}{|l|}{PSPNet \cite{pspnet}} &
  \multicolumn{1}{l|}{ResNet50} &
  \multicolumn{1}{l|}{0.83} &
  \multicolumn{1}{l|}{0.08} &
  \multicolumn{1}{l|}{0.90} &
  \multicolumn{1}{l|}{0.07} \\ \hline
\multicolumn{1}{|l|}{SegNet \cite{segnet}} &
  \multicolumn{1}{l|}{VGG16} &
  \multicolumn{1}{l|}{0.82} &
  \multicolumn{1}{l|}{0.09} &
  \multicolumn{1}{l|}{0.90} &
  \multicolumn{1}{l|}{0.05} \\ \hline
\multicolumn{1}{|l|}{UNet \cite{unet}} &
  \multicolumn{1}{l|}{VGG16} &
  \multicolumn{1}{l|}{0.5} &
  \multicolumn{1}{l|}{0.26} &
  \multicolumn{1}{l|}{0.67} &
  \multicolumn{1}{l|}{0.15} \\ \hline
\multicolumn{1}{|l|}{FCN32 \cite{fcn32}} &
  \multicolumn{1}{l|}{VGG16} &
  \multicolumn{1}{l|}{0.76} &
  \multicolumn{1}{l|}{0.13} &
  \multicolumn{1}{l|}{0.91} &
  \multicolumn{1}{l|}{0.02} \\ \hline
\multicolumn{1}{|l|}{PSPNet \cite{pspnet}} &
  \multicolumn{1}{l|}{VGG16} &
  \multicolumn{1}{l|}{0.82} &
  \multicolumn{1}{l|}{0.09} &
  \multicolumn{1}{l|}{0.92} &
  \multicolumn{1}{l|}{0.02} \\ \hline
\multicolumn{1}{|l|}{\textbf{FloodTransformer (Ours)}} &
  \multicolumn{1}{l|}{Transformer} &
  \multicolumn{1}{l|}{\textbf{0.93}} &
  \multicolumn{1}{l|}{\textbf{0.03}} &
  \multicolumn{1}{l|}{\textbf{0.96}} &
  \multicolumn{1}{l|}{\textbf{0.02}} \\ \hline
 &
   &
   &
   &
   &
   \\
 &
   &
   &
   &
   &
   \\

\end{tabular}
\vspace{-20pt}
\caption{\centering Average and Standard deviation of Mean Intersection over Union
(mIoU) and Pixel Accuracy (PA) on WSOC dataset}
\label{tab:my-table-1}
\end{table*}

\section{Conclusions and Impact}
In this work we propose, to the best of our knowledge, the first hybrid transformer-CNN model for flood water segmentation: \textit{FloodTransformer}. We achieve state-of-the-art segmentation results on the WSOC dataset and empirically demonstrate the power and robustness of visual transformers, deviating from pure CNN-based approaches.

We further validate our model's performance by demonstrating competitive qualitiative results on out-of-dataset aerial images, indicating competitive performance on images in the wild. With the increasing number and variability of flooding-related disasters in developing countries (varying weather conditions, water flow, geography etc.) and variability of visual data collected from UAVs (varying mode of capture, camera technology etc.), this generalization capacity is crucial. 

The flood water segmentation methodology adapted in this paper is useful for Flood Prediction and Response. Given the high incidence of flooding-related disasters, our method is an important step towards improving flood warning systems and covering billions of people in high-risk developing countries. It can also aid in the first step of disaster management: Post Disaster Needs Assessment (PDSA). In particular, in both cases it reduces the need for manual analysis; this is crucial in developing countries with constrained resources and high populations, where speed and efficiency of response is of paramount importance. 





\appendix


\newpage
{\small
\bibliographystyle{plain}
\bibliography{neurips_2021}
}

\end{document}